\documentclass[10pt,letterpaper]{article}

\usepackage[T1]{fontenc}
\usepackage[utf8]{inputenc}
\usepackage{mathptmx}                 
\usepackage{amsmath,amssymb,bm}
\usepackage[letterpaper,margin=1in,headheight=14pt]{geometry}
\usepackage{graphicx}
\usepackage{booktabs}
\usepackage{xcolor}
\usepackage{wrapfig}
\usepackage{enumitem}
\usepackage{microtype}
\usepackage{xspace}
\usepackage{titlesec}
\usepackage{fancyhdr}
\usepackage[most]{tcolorbox}
\usepackage[font=small,labelfont={bf,color=splatmagenta},labelsep=period]{caption}
\usepackage[numbers,sort&compress]{natbib}

\definecolor{splatmagenta}{HTML}{C2407E}
\definecolor{splatmagentalight}{HTML}{FBE7F0}
\definecolor{splatteal}{HTML}{22A5BB}
\definecolor{splatorange}{HTML}{EB6834}
\definecolor{splatgray}{HTML}{52514E}

\newcommand{\displayfont}{\fontfamily{pag}\selectfont}

\titleformat{\section}{\displayfont\Large\bfseries\color{splatmagenta}}{\thesection}{0.6em}{}
\titleformat{\subsection}{\displayfont\large\bfseries\color{splatmagenta}}{\thesubsection}{0.6em}{}
\titleformat{\subsubsection}{\displayfont\normalsize\bfseries\color{splatmagenta}}{\thesubsubsection}{0.6em}{}
\titleformat{\paragraph}[runin]{\displayfont\normalsize\bfseries\color{splatgray}}{}{0pt}{}[.]
\titlespacing*{\section}{0pt}{1.6ex plus .4ex minus .2ex}{0.8ex plus .2ex}
\titlespacing*{\subsection}{0pt}{1.3ex plus .3ex minus .2ex}{0.6ex plus .2ex}
\titlespacing*{\paragraph}{0pt}{0.9ex plus .3ex minus .2ex}{0.6em}

\renewcommand{\headrulewidth}{0.4pt}
\renewcommand{\headrule}{\hbox to\headwidth{\color{splatmagenta}\leaders\hrule height \headrulewidth\hfill}}
\fancypagestyle{plain}{\fancyhf{}\renewcommand{\headrulewidth}{0pt}\fancyfoot[C]{\displayfont\small\color{splatgray}\thepage}}

\setlist{itemsep=0.2em,topsep=0.3em}

\usepackage[breaklinks,colorlinks,allcolors=splatmagenta]{hyperref}
\usepackage{cleveref}
\crefname{section}{Sec.}{Secs.}
\crefname{subsection}{Sec.}{Secs.}
\crefname{appendix}{App.}{Apps.}
\crefname{figure}{Fig.}{Figs.}
\crefname{table}{Tab.}{Tabs.}
\crefname{equation}{Eq.}{Eqs.}

\newcommand{\splattok}{SplatTok\xspace}

\newcommand{\reporttitle}{SplatTok: Data-Driven Compression of 3D Gaussian Splats with Scene-Level Tokens}
\newcommand{\reportauthors}{Sai Karthikey Pentapati$^{1}$, Shashank N. Sridhara$^{2}$, Birendra Kathariya$^{2}$, Peng Yin$^{2}$, Guandao Yang$^{1}$, Alan C. Bovik$^{3}$}
\newcommand{\reportaffiliations}{$^{1}$The University of Texas at Austin \quad $^{2}$Dolby Laboratories Inc. \quad $^{3}$University of Colorado Boulder}

\newcommand{\reportcorrespondence}{karthikey97@gmail.com}
\newcommand{\reportprojectpage}{https://karthikey97.github.io/splattok/}

\begin{document}

\thispagestyle{plain}
\begin{center}
  {\displayfont\bfseries\LARGE\color{splatmagenta}\reporttitle\par}
  \vspace{1.0em}
  {\displayfont\large \reportauthors\par}
  \vspace{0.4em}
  {\displayfont\small\color{splatgray}\reportaffiliations\par}
  \vspace{0.6em}
  {\displayfont\small\color{splatgray} Correspondence: \href{mailto:\reportcorrespondence}{\reportcorrespondence}\quad\textbar\quad Project page: \url{\reportprojectpage}\par}
\end{center}
\vspace{0.6em}

\begin{tcolorbox}[enhanced,colback=splatmagentalight,colframe=splatmagentalight,boxrule=0pt,arc=2pt,
  left=10pt,right=10pt,top=8pt,bottom=8pt,borderline west={2.5pt}{0pt}{splatmagenta}]
{\displayfont\bfseries\color{splatmagenta}Abstract}\par\vspace{0.3em}
3D Gaussian splats enable real-time rendering but require substantial storage.
Existing compression methods either encode the primitives of a given splat or optimize a compact representation independently for each scene.
Yet a Gaussian splat is only one realization of a scene: different arrangements of primitives can produce nearly identical rendered views.
We exploit this freedom on the decoding side with SplatTok, a learned scene-resynthesis codec.
Trained across thousands of real-world scenes, a transformer encodes an existing splat into compact, spatially anchored latent tokens, and a shared decoder synthesizes a new set of standard Gaussian primitives without preserving correspondence to the input.
The representation is learned through rendering supervision rather than primitive reconstruction.
At compression time, a quantization-aware latent search refines only the scene code, keeping the networks frozen, and closes most of the encoder's remaining error.
On standard benchmarks, using only renders of the input splat for the latent search, without the original training images, SplatTok matches the PSNR of the feed-forward codec FCGS with 4 to 6.5 times smaller files, compressing input splats about 75 to 110 times.
With the training images, it exceeds the PSNR of the per-scene method HAC++ with 13 to 41\% smaller files.
A single model trained on DL3DV generalizes to unseen scenes with millions of Gaussians and supports multiple target rates without retraining.

\end{tcolorbox}

\vspace{0.4em}
\begin{figure}[h]
  \centering
  \includegraphics[width=\textwidth]{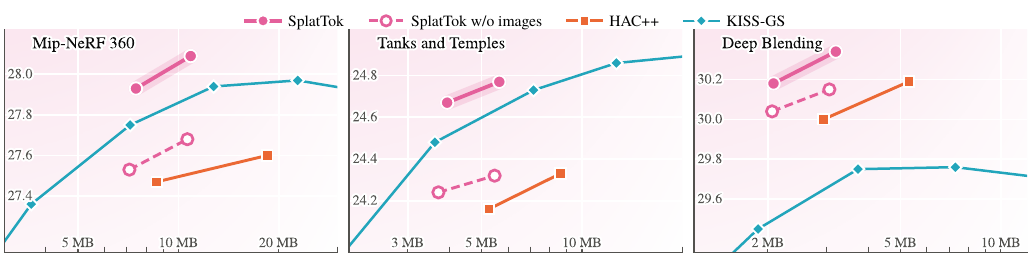}
  \caption{\textbf{Compressing an existing Gaussian splat.}
  PSNR (dB) against size per scene (MB, log scale).
  With the training images as the target of its latent search, \splattok outperforms HAC++~\cite{chen2025hacpp} and KISS-GS~\cite{morgenstern2026kissgs} in PSNR on all three datasets.
  With only renders of the input splat as the target, it still lies above HAC++, which is trained on the training images.
  All \splattok points compress the 3DGS-MCMC splats of \cref{tab:gt}.}
  \label{fig:teaser}
\end{figure}

\section{Introduction}
\label{sec:intro}

3D Gaussian Splatting~\cite{kerbl20233d} has quickly become a standard way to capture and share photorealistic 3D scenes.
A scene is reconstructed once from multi-view images captured from many camera poses and is then rendered in real time on ordinary hardware.
Rendering is fast because the scene is stored directly as millions of Gaussian primitives, each with dozens of attributes, and that is also what makes it large: a typical scene occupies hundreds of megabytes to over a gigabyte.
That is too large to stream to a headset or phone, to store by the thousands in an asset library, or to send as easily as an image or a video.
In this work we compress existing splats.
We want a file one to two orders of magnitude smaller that renders nearly as well as the original, produced quickly, without the images the splat was built from, and that decodes back to a standard splat any 3DGS renderer can display.
The last two conditions matter in practice: splats are increasingly distributed as standalone files, long after their training images have been discarded, and they are viewed in renderers built for the standard format.

Existing compression methods each meet some of these goals but not all of them.
The most compact results come from per-scene methods, which build a new, structured representation for every scene and optimize it against the scene's training images, often with learned context models for entropy coding~\cite{lu2024scaffold,chen2024hac,chen2025hacpp,wang2024contextgs}; see~\cite{bagdasarian2025survey} for a survey.
The strongest of these reach compression ratios near 100-fold, but they need the training images, take tens of minutes per scene, and usually must be retrained for every target size.
Their output is also not a standard splat: anchor-based methods store learned features from which small networks generate the Gaussians for each view at render time, so they need their own renderer.
Post-training methods prune and quantize a splat without building a new representation~\cite{fan2024lightgaussian,lee2024compact,niedermayr2024compressed,xie2024mesongs,tian2025flexgaussian} and keep the standard format, but reach more modest ratios or need minutes of fine-tuning to recover quality.
The feed-forward codec FCGS~\cite{chen2025fcgs} removes per-scene optimization altogether: a network trained on many scenes compresses a splat in seconds, without images, and decodes to a standard splat.
It does so primitive by primitive, coding each Gaussian's attributes with limited context from its neighbors and keeping the original set of Gaussians, and reaches about 10 to 30-fold compression.
No existing method reaches per-scene compression ratios without the training images and without per-scene representation training, while decoding to a standard splat.

We take a different view: rather than learning a new representation for each scene, or a codec for individual Gaussians, we learn a model of whole scenes and compress within it.
Our method, \splattok, encodes an entire splat into a compact set of latent tokens anchored at 3D locations across the scene, and decodes them into a new set of Gaussians placed around those anchors.
The encoder attends across the whole scene, so each token summarizes its region in the context of everything else; the decoder never sees the input Gaussians and is free to synthesize whatever arrangement renders the scene best.
We train the model on a large dataset of real-world scenes, supervised only by rendering, and quantize and entropy code the tokens to form the bitstream.
This differs from FCGS in what is learned and what is stored: FCGS learns to code the attributes of given primitives, whereas \splattok learns how scenes are structured and stores a description from which primitives are regenerated.

Because the decoder maps any code in its latent space to a renderable splat, compressing a new scene reduces to finding its code.
The encoder's output is already a good latent code, and a short search over the tokens alone, with the network frozen and quantization applied during the search, brings it close to the best code the decoder can express.
The search needs only renders of the input splat from its cameras as its target, so it works without the training images; when they are available, they can serve as the target instead.
This mirrors encode-time refinement in learned image compression~\cite{campos2019content,yang2020improving}, here applied to whole 3D scenes.
Our analysis shows why it is effective: the search changes the code by only 4 to 10\% yet improves fidelity to the input splat by about 8\,dB, indicating that refinement closes at least 84\% of the initial optimality gap within the fixed decoder and admissible code space.
The rate is set by the precision of the tokens and by the density of the anchors, so a single trained model serves many target sizes, and the same weights are used for every dataset in our experiments.

The resulting codec outperforms the best published per-scene methods in PSNR at smaller sizes.
On Mip-NeRF 360~\cite{barron2022mipnerf360}, \splattok reaches 27.93\,dB at 7.46\,MB per scene, 0.46\,dB above HAC++~\cite{chen2025hacpp} at 8.61\,MB, and 28.09\,dB at 10.89\,MB, 0.49\,dB above HAC++ at 18.48\,MB, with similar gains on Tanks and Temples~\cite{knapitsch2017tanks} and Deep Blending~\cite{hedman2018deep}, using one model for all sizes instead of one training run per size.
With coarser anchors, the same model still exceeds HAC++ in PSNR with files 53 to 60\% smaller.
Without the training images, \splattok matches the PSNR of FCGS on the same input splats with 4 to 6.5 times smaller files on DL3DV, Mip-NeRF 360, Tanks and Temples and Deep Blending.
The same model, trained only on DL3DV splats, compresses scenes of several million Gaussians from these other datasets without retraining.

In summary, our contributions are:
\begin{itemize}
  \item A scene-resynthesis codec for Gaussian splats that encodes a whole scene into a compact set of 3D-anchored tokens and synthesizes a new set of standard 3DGS primitives from them, trained across thousands of real scenes with rendering supervision alone.
  \item Compression as a short search in the decoder's latent space, which reaches per-scene compression quality without the training images, without per-scene representation training, and with one model for all target sizes. We show that it works because most of the encoder's remaining error is an amortization gap rather than a limit of the decoder.
  \item Results on DL3DV, Mip-NeRF 360, Tanks and Temples and Deep Blending: higher PSNR than HAC++ with 13 to 41\% smaller files at the default anchor density and 53 to 60\% smaller files with coarser anchors, the PSNR of FCGS with 4 to 6.5 times smaller files without training images, and generalization from 350,000 to several million Gaussians per scene.
\end{itemize}

The rest of this report is organized as follows.
\Cref{sec:related} reviews related work, \cref{sec:background} fixes notation, and \cref{sec:method} describes \splattok.
\Cref{sec:experiments} evaluates it against feed-forward, post-training and per-scene methods, studies its rate controls, the latent search and its design choices, and reports its timing.
\Cref{sec:discussion} discusses limitations and future directions.

\section{Related Work}
\label{sec:related}

\paragraph{Compact Gaussian splats trained per scene}
A first line of work makes the splat itself smaller while it is fit to the training images.
These methods learn masks that prune redundant Gaussians~\cite{lee2024compact,papantonakis2024reduced}, replace attributes with codebook indices or quantized latents~\cite{lee2024compact,navaneet2024compgs,girish2024eagles}, arrange attributes into smooth 2D grids that standard image codecs compress well~\cite{morgenstern2024sog}, or add an explicit rate term to the training objective~\cite{wang2024rdogaussian}.
SpeedyGS~\cite{speedygs} compresses ordinary 3DGS primitives with a content-aware two-stage optimization.
These methods keep the output close to a standard splat and reach roughly 15 to 45-fold compression on Mip-NeRF 360, but each scene still needs a full optimization from its training images, and changing the target size means optimizing again.

\paragraph{Anchor-based methods with learned entropy models}
The smallest files come from methods that replace the unstructured splat with a structured one.
Scaffold-GS~\cite{lu2024scaffold} stores sparse anchors with learned features, from which small networks generate nearby Gaussians for each view.
HAC~\cite{chen2024hac} and HAC++~\cite{chen2025hacpp} model the distribution of these anchor features with a hash-grid context and, in HAC++, context within each anchor, while ContextGS~\cite{wang2024contextgs} decodes anchors coarse to fine with each level conditioned on the previous ones, and CompGS~\cite{liu2024compgs} predicts primitives as residuals from anchors.
Recent methods refine the context model further, with context-adaptive triplanes~\cite{zhan2025cat3dgs}, hierarchical sparse transform coding~\cite{shtc}, local geometry-aware hierarchical context~\cite{lghcc}, or anchor-wise causal context with integer-only inference~\cite{cosags}.
Together they reach close to 100-fold compression, the best reported.
The price is a per-scene optimization of tens of minutes against the training images, repeated for every target rate except in progressive coding~\cite{chen2026pcgs}, which trains one representation whose bitstream can be decoded at several rates, and an output that is not a standard splat but needs the method's own networks at render time.
\splattok instead learns the structure of scenes once, across thousands of them, so that a new scene needs no representation of its own, and its latents decode to ordinary Gaussians.

\paragraph{Post-training compression}
Another family compresses an already trained splat.
LightGaussian~\cite{fan2024lightgaussian} prunes Gaussians by their contribution to the training views, distills spherical harmonics to a lower degree and vector-quantizes the rest; Compressed3D~\cite{niedermayr2024compressed} clusters attributes by their sensitivity to rendering error; MesonGS~\cite{xie2024mesongs} combines importance-based pruning with octree coding of positions and a transform coder for attributes.
Each then fine-tunes the compressed splat against the training images for several minutes.
FlexGaussian~\cite{tian2025flexgaussian} removes the fine-tuning and chooses pruning and mixed-precision quantization without training.
KISS-GS~\cite{morgenstern2026kissgs} combines strong compaction with an image-based attribute codec and encoding-aware fine-tuning, and reaches the highest ratios of this family while keeping a standard splat.
These methods keep the standard format, but reach more modest ratios, especially without fine-tuning.
\splattok's latent search instead optimizes the latent code itself, with quantization in the loop, while the frozen decoder, trained across thousands of scenes, keeps the code within the space of plausible scenes.
This reaches several times higher compression ratios than pruning and quantizing primitives, and since \splattok accepts any standard splat, it is complementary to these methods and can further compress a splat they have already compacted (\cref{subsec:rate}).

\paragraph{Feed-forward compression}
FCGS~\cite{chen2025fcgs} is, to our knowledge, the first codec that compresses a splat in a single forward pass of a network trained on many scenes.
It needs no images and no per-scene optimization, decodes to a standard splat, and runs in seconds.
It codes the given Gaussians themselves: each primitive's attributes are entropy coded with context from already decoded neighbors, and the decoded splat has the same set of Gaussians as the input, which limits compression to about 10 to 30-fold.
LocoMoco~\cite{locomoco} extends FCGS with long-context entropy modeling over Morton-ordered windows of Gaussians, and still codes the given primitives; without public code or models, we do not include it in our comparisons.
Feed-forward reconstruction methods have also been combined with compression of their predicted Gaussians~\cite{gensplatcodec,codecsplat}; these compress the output of reconstruction from images rather than an existing splat.
\splattok is also trained across scenes, but it learns a model of whole scenes rather than of individual Gaussians and stores a compact description from which a new set of Gaussians is generated.

\paragraph{Learned compression and encode-time refinement}
Learned image codecs pair analysis and synthesis transforms with an entropy model~\cite{balle2017end,balle2018variational,minnen2018joint}.
Since an amortized encoder does not find the best code for every input, refining each latent at encode time with the decoder fixed recovers part of this gap~\cite{campos2019content,yang2020improving}.
Implicit neural representations take the opposite extreme and overfit a network to each instance~\cite{dupont2021coin}, which is also what per-scene splat compression methods do.
\splattok sits between the two: a codec amortized over thousands of scenes and refined per scene at encode time with the network frozen, and we show that this refinement closes most of the gap to the best code the decoder can express.

\paragraph{Set latents and latent models of Gaussian splats}
Our architecture builds on set latents for 3D shapes, which encode a point cloud into a fixed-size set of latent vectors through cross-attention~\cite{zhang20233dshape2vecset} and now underlie many 3D generative models~\cite{zhang2024clay}.
Latent models of Gaussian splats~\cite{roessle2024l3dg,DeG} are built to support a generative prior, mostly at object level, rather than to reconstruct a given scene faithfully, and are not evaluated in terms of size and rendering quality.
At scene level, Can3Tok~\cite{gao2025can3tok} is a variational autoencoder for 3D Gaussians, designed to provide a latent space for diffusion-based scene generation.
A fixed set of learned queries cross-attends to the input Gaussians, so every scene, whatever its extent or complexity, is mapped to a latent of the same size, and the decoder always outputs the same number of Gaussians.
To fit this budget, each input splat is first filtered and subsampled to a fixed number of primitives (40K in their experiments), which discards most of a real capture.
The model is trained with an L2 loss on the attributes of these input Gaussians rather than on renderings.
Since many different sets of Gaussians render the same scene, this loss asks the decoder to reproduce one particular arrangement of primitives rather than the scene's appearance, and reconstructions are measured by attribute error rather than rendering fidelity.
Its checkpoints are not publicly available, which prevents a direct comparison.
GaussianGPT~\cite{lutzow2026gaussiangpt} generates scenes autoregressively over the discrete tokens of a sparse 3D convolutional autoencoder with lookup-free quantization.
Its encoder takes a voxel grid with at most one Gaussian per voxel, stored as an offset from the voxel center; when a voxel contains several Gaussians, all but one are randomly dropped.
Its training scenes are therefore built to match this structure, by fitting Scaffold-GS with one Gaussian per anchor aligned to the voxel centers, and view-dependent color is not modeled for scenes.
A splat trained with standard 3DGS, with many overlapping primitives of different sizes per voxel, falls outside this input domain.
\splattok instead grows the number of anchors, and hence the code and the output, with the input splat, accepts any standard splat, and is supervised only through rendering.

\paragraph{Scene tokens from images}
TokenGS~\cite{ren2026tokengs} and SceneTok~\cite{asim2026scenetok} also describe a scene with a set of learned tokens, but are built for feed-forward reconstruction from posed images rather than for compressing a splat.
In TokenGS, learnable 3DGS tokens cross-attend to features of a few input views and each regresses a fixed number of Gaussians, trained with a rendering loss.
SceneTok encodes many context views into a small set of unstructured tokens, from which a rectified-flow decoder generates novel views as images rather than an explicit 3D representation.
Neither takes an existing splat as input or produces a bitstream evaluated in size and rendering fidelity; their architectures, losses and pipelines target the fast construction of renderable scenes from images.
\splattok instead starts from a finished splat, anchors its tokens in 3D across the whole scene, and is trained and evaluated as a codec.

\section{Background and Notation}
\label{sec:background}

A Gaussian splat~\cite{kerbl20233d} represents a scene as a set of $N$ anisotropic 3D Gaussians $\mathcal{G}=\{g_i\}_{i=1}^{N}$.
Each Gaussian $g_i=(\boldsymbol{\mu}_i,\mathbf{s}_i,\mathbf{q}_i,o_i,\mathbf{f}_i)$ has a position $\boldsymbol{\mu}_i\in\mathbb{R}^3$, a log-scale $\mathbf{s}_i\in\mathbb{R}^3$, a rotation given by a quaternion $\mathbf{q}_i\in\mathbb{R}^4$, an opacity logit $o_i\in\mathbb{R}$, and spherical-harmonic (SH) color coefficients $\mathbf{f}_i\in\mathbb{R}^{(\ell+1)^2\times 3}$ up to degree $\ell$.
Its covariance is $\boldsymbol{\Sigma}_i=\mathbf{R}(\mathbf{q}_i)\,\mathrm{diag}(e^{\mathbf{s}_i})^2\,\mathbf{R}(\mathbf{q}_i)^{\top}$ and its opacity is $\mathrm{sigmoid}(o_i)$.
We write $\mathcal{R}(\mathcal{G},\mathbf{P})$ for the image of a splat rendered from a camera $\mathbf{P}$, and treat the renderer as a fixed, differentiable operator.
A splat is fit to multi-view images $\{\mathbf{I}_v\}_{v\in\mathcal{V}}$, the training images, taken from cameras $\{\mathbf{P}_v\}_{v\in\mathcal{V}}$, where $\mathcal{V}=\{1,\dots,V\}$ indexes the views.

\newpage
\section{SplatTok}
\label{sec:method}

\begin{figure}[t]
  \centering
  \includegraphics[width=0.9\textwidth]{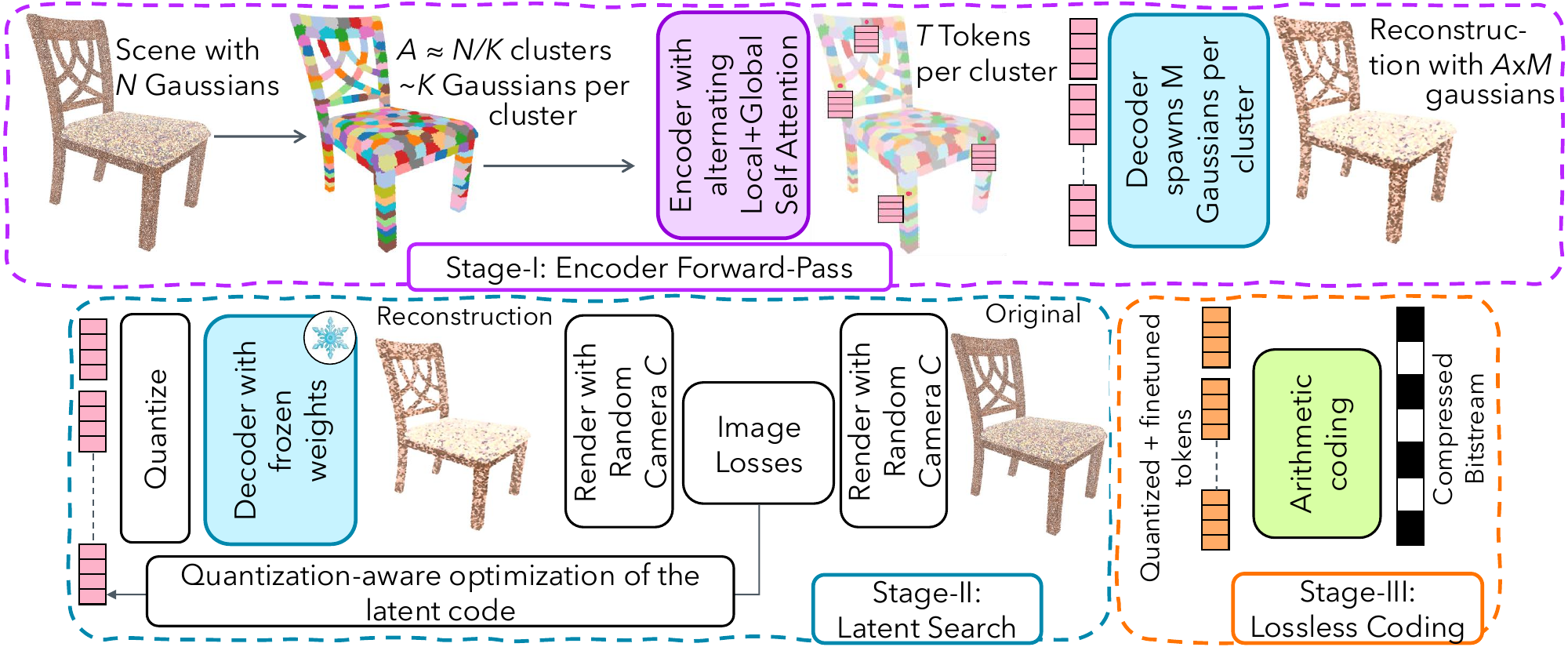}
  \caption{\textbf{Overview of \splattok.}
  The input splat is partitioned into anchors.
  The encoder summarizes the Gaussians of each anchor into a fixed number of latent tokens.
  The decoder regenerates a fixed number of new Gaussians around each anchor from its tokens.
  At encode time, latent search refines the quantized tokens with both networks frozen.}
  \label{fig:pipeline}
\end{figure}

\subsection{Overview}
\label{subsec:overview}

\splattok compresses a splat $\mathcal{G}$ into a set of latent tokens anchored at 3D locations and regenerates a new splat $\hat{\mathcal{G}}$ from them (\cref{fig:pipeline}).
We first partition the scene into $A$ anchors, each with a position and a size (\cref{subsec:anchors}).
An encoder $E$ turns the Gaussians around each anchor into $T$ tokens of $C$ channels, attending across the whole scene, which gives the encoder's code $\mathbf{Z}^0=E(\mathcal{G})\in\mathbb{R}^{A\times T\times C}$ (\cref{subsec:encoder}).
A decoder $D$ maps the tokens of each anchor of a code $\mathbf{Z}\in\mathbb{R}^{A\times T\times C}$ to $M$ new Gaussians placed around it, so that $\hat{\mathcal{G}}=D(\mathbf{Z})$ is a standard splat of $AM$ Gaussians that any 3DGS renderer can display (\cref{subsec:decoder}).
Here, hats mark reconstructed quantities.
Both networks are trained jointly across many scenes through rendering alone (\cref{subsec:training}).
To store a scene, the tokens are quantized to $b$ bits and entropy coded (\cref{subsec:coding}).
Finally, a latent search refines the quantized code at encode time with both networks frozen, using renders of the input splat or, when available, the training images as the target (\cref{subsec:search}).

\paragraph{Input and normalization}
\splattok takes a splat exactly as it is stored: an unordered set of Gaussians in the parametrization of \cref{sec:background}, of any size $N$, without the training images.
Before encoding, the splat and its cameras are brought to a normalized frame by a similarity transform computed from the cameras, consistent with the normalization of the \textit{gsplat} library~\cite{ye2024gsplat}, so that splats of different scenes share a common position and scale.
The decoded splat lives in the same frame and is rendered with the transformed cameras.

\subsection{Scene partition into anchors}
\label{subsec:anchors}

We partition the scene into $A$ anchors, chosen proportional to $N$ for a fixed target number of Gaussians per anchor, so that the number of anchors follows the size of the scene.
The anchor positions $\{\mathbf{a}_k\}_{k=1}^{A}$ are the centers of a $k$-means clustering of all Gaussian positions $\{\boldsymbol{\mu}_i\}$.
Clusters with more than $K$ members are split.
Each Gaussian joins its nearest anchor and, as long as that anchor has fewer than $K$ members, also its second-nearest one.
The resulting member sets $\mathcal{N}_k$, with $|\mathcal{N}_k|\le K$, overlap near cluster boundaries, so that neighboring anchors see the Gaussians between them.
Each anchor also stores a size $\sigma_k$, the median distance of its members from $\mathbf{a}_k$, capped at a multiple of the scene's median anchor size so that isolated anchors on sparse outliers do not span large regions.
Positions inside an anchor are expressed as normalized offsets $(\boldsymbol{\mu}_i-\mathbf{a}_k)/\sigma_k$, so that every anchor is processed at a common scale, whether it covers a dense surface or a sparse background region.
The anchor positions and sizes are the only geometric side information of the code.

\subsection{Encoder}
\label{subsec:encoder}

Although anchors may contain different numbers of Gaussians, the encoder summarizes every anchor into the same number $T$ of tokens.
Each Gaussian is first embedded from its non-positional attributes, $\mathbf{e}_i=\phi(\mathbf{s}_i,\bar{\mathbf{q}}_i,o_i,\mathbf{f}_i)$, where $\phi$ is an MLP and $\bar{\mathbf{q}}_i$ is the unit quaternion.
Within anchor $k$, the embedding is combined with the member's normalized offset into a key--value token
\begin{equation}
  \mathbf{h}_{k,i}=\psi\big(\mathbf{e}_i,\ \gamma\big((\boldsymbol{\mu}_i-\mathbf{a}_k)/\sigma_k\big)\big),\qquad i\in\mathcal{N}_k,
  \label{eq:kv}
\end{equation}
where $\gamma$ is a Fourier embedding and $\psi$ an MLP.
A set of $T$ learned queries $\{\mathbf{w}^{\mathrm{e}}_t\}_{t=1}^{T}$, shared by all anchors and shifted by an embedding of the anchor size, $\mathbf{w}^{\mathrm{e}}_t+\mathbf{W}^{\mathrm{e}}\log\sigma_k$, cross-attends to $\{\mathbf{h}_{k,i}\}_{i\in\mathcal{N}_k}$, followed by a feed-forward layer.
Because attention normalizes over however many members an anchor has, the output is always $T$ tokens per anchor.

This is the central structural choice of \splattok.
The code length of an anchor is fixed, independent of how many Gaussians fall into it, how they are arranged, and what their individual attributes are.
The bitstream therefore grows with the number of anchors and the token precision, and the decoder is free to represent an anchor with a different set of Gaussians than the input had.
FCGS~\cite{chen2025fcgs}, by contrast, codes each given Gaussian, so its code length is tied to the input primitives.
It also makes the anchor density a rate control: placing fewer anchors per scene shortens the code without retraining (\cref{subsec:rate}).

The tokens of all anchors are then processed jointly by a transformer, so that each token summarizes its region in the context of the rest of the scene.
Before the transformer, a learned embedding of the token index $t$ is added to each token.
The first few layers apply self-attention among the $T$ tokens of each anchor, and the remaining layers alternate between this local attention and global self-attention over all $AT$ tokens of the scene, similar to VGGT~\cite{vggt}.
In the global layers, queries and keys carry a 3D rotary position encoding~\cite{su2024roformer} of their anchor position $\mathbf{a}_k$, shared by the $T$ tokens of the anchor, so that attention between tokens depends on the relative position of their anchors.
Global attention gives each token context beyond its own anchor, which a per-anchor or per-primitive encoder cannot use; restricting the encoder to attention within anchors lowers the quality of the reconstructed scene (\cref{subsec:ablations}).
A final layer normalization and linear projection reduce each token to $C$ channels, giving the tokens $\mathbf{z}^0_{k,t}\in\mathbb{R}^{C}$ of the encoder's code $\mathbf{Z}^0$.

\subsection{Decoder}
\label{subsec:decoder}

The decoder generates the Gaussians of each anchor from that anchor's tokens alone.
A set of $M$ learned queries $\{\mathbf{w}^{\mathrm{d}}_m\}_{m=1}^{M}$, shared by all anchors and shifted by a size embedding $\mathbf{W}^{\mathrm{d}}\log\sigma_k$, cross-attends to the anchor's tokens $\{\mathbf{z}_{k,t}\}_{t=1}^{T}$ of a code $\mathbf{Z}$, and a residual MLP maps each attended query to the parameters $(\boldsymbol{\delta}_{k,m},\hat{\mathbf{s}}_{k,m},\hat{\mathbf{q}}_{k,m},\hat{o}_{k,m},\hat{\mathbf{f}}_{k,m})$ of one Gaussian.
The Gaussian is placed relative to its anchor,
\begin{equation}
  \hat{\boldsymbol{\mu}}_{k,m}=\mathbf{a}_k+\sigma_k\,\boldsymbol{\delta}_{k,m},
  \label{eq:offset}
\end{equation}
and its log-scale, rotation and opacity logit are used directly in the parametrization of \cref{sec:background}.
Color is encoded and predicted in a luma--chroma (BT.709 YCbCr) SH basis in which only luma carries view-dependent terms and the two chroma channels keep only the DC term.
This removes most color coefficients at a cost of less than 0.1\,dB on the input splats, and the predicted colors are converted back to RGB SH.
The decoded splat $\hat{\mathcal{G}}=D(\mathbf{Z})=\{\hat g_{k,m}\}$ has $AM$ Gaussians in the standard format.
The decoder never sees the input Gaussians, so it is not constrained to reproduce them: it may place, size and weight its primitives differently from the input wherever that renders the scene better.
Each anchor is decoded independently, which keeps decoding parallel and its memory proportional to the number of anchors.

\subsection{Training}
\label{subsec:training}

The encoder and decoder are trained jointly across many scenes, supervised only by rendering.
For a training scene, we encode and decode its splat and render the result from a random subset $\mathcal{V}'\subset\mathcal{V}$ of its cameras, drawn anew at every step.
Each render is compared with a target image $\mathbf{Y}_v$; for a rendered image $\hat{\mathbf{I}}$ and a target $\mathbf{Y}$, the loss is
\begin{equation}
  \mathcal{L}(\hat{\mathbf{I}},\mathbf{Y})=(1-\lambda)\,\mathrm{MSE}(\hat{\mathbf{I}},\mathbf{Y})+\lambda\,\big(1-\mathrm{SSIM}(\hat{\mathbf{I}},\mathbf{Y})\big),
  \label{eq:loss}
\end{equation}
where $\mathrm{MSE}$ is the mean squared error over pixels.
The training objective is
\begin{equation}
  \mathbb{E}_{\mathcal{G},\,\mathcal{V}'}\Big[\frac{1}{|\mathcal{V}'|}\sum_{v\in\mathcal{V}'}\mathcal{L}\big(\mathcal{R}(D(E(\mathcal{G})),\mathbf{P}_v),\ \mathbf{Y}_v\big)\Big],
  \label{eq:train}
\end{equation}
with the expectation over training scenes and view subsets.
During training the targets are the training images, $\mathbf{Y}_v=\mathbf{I}_v$.
There is no loss on the attributes of individual Gaussians, and no correspondence between input and output primitives is needed, since the decoder generates a different set.
The networks are trained on continuous tokens; quantization is applied only when a scene is stored (\cref{subsec:coding}).

\subsection{Quantization and entropy coding}
\label{subsec:coding}

To store a scene, the tokens are quantized channel-wise with a symmetric uniform quantizer of $b$ bits.
With $u_{\max}=2^{b-1}-1$ and a per-channel step $\Delta_c=\max_{k,t}|z^{0}_{k,t,c}|/u_{\max}$ set from the encoder's code $\mathbf{Z}^0$, each value of a code $\mathbf{Z}$ is mapped to an integer symbol and back,
\begin{equation}
  u_{k,t,c}=\mathrm{round}\big(\mathrm{clip}(z_{k,t,c}/\Delta_c,\,-u_{\max},\,u_{\max})\big),\qquad
  \hat z_{k,t,c}=\Delta_c\,u_{k,t,c},
  \label{eq:quant}
\end{equation}
and we write $\hat{\mathbf{Z}}=Q_b(\mathbf{Z})$ for the dequantized code.

The symbols are coded losslessly with rANS~\cite{duda2013asymmetric}.
Anchors are coded in the Morton order of their positions.
The probability table of each symbol $u_{k,t,c}$ is selected by its position $(t,c)$ within the anchor and by a coarse bin of the symbol at the same $(t,c)$ in $\pi(k)$, the nearest anchor coded before $k$.
The tables are integer frequency counts, collected once for each bit depth from encoder codes of training scenes, and are fixed and shared by encoder and decoder: like the network weights, they are part of the codec rather than of any bitstream, and there is no learned entropy model.
Besides the coded symbols, the bitstream holds the anchor positions $\mathbf{a}_k$, the anchor sizes $\sigma_k$ and the per-channel steps $\Delta_c$.
Without the latent search, the stored code is $\hat{\mathbf{Z}}^0=Q_b(\mathbf{Z}^0)$; the search of \cref{subsec:search} replaces it with a refined code at the same bit depth.
The bit depth $b$ sets the rate, so a single trained model serves different target sizes.

\subsection{Latent search}
\label{subsec:search}

The encoder is amortized over many scenes and does not find the best code for each one.
At encode time, we therefore refine the code with both encoder and decoder frozen, as in encode-time refinement for learned image compression~\cite{campos2019content,yang2020improving}.
Starting from $\mathbf{Z}^0$, we solve
\begin{equation}
  \mathbf{Z}^{\star}=\arg\min_{\mathbf{Z}}\ \frac{1}{|\mathcal{V}|}\sum_{v\in\mathcal{V}}\mathcal{L}\big(\mathcal{R}(D(\hat{\mathbf{Z}}),\mathbf{P}_v),\,\mathbf{Y}_v\big),\qquad
  \hat{\mathbf{Z}}=Q_b(\mathbf{Z}),
  \label{eq:search}
\end{equation}
by stochastic gradient descent, with a random subset $\mathcal{V}'\subset\mathcal{V}$ of the cameras at each step, and store $\hat{\mathbf{Z}}^{\star}=Q_b(\mathbf{Z}^{\star})$.
The network weights, the anchors $\mathbf{a}_k$, the sizes $\sigma_k$ and the steps $\Delta_c$ stay fixed, so only the coded symbols change.
Quantization is applied inside the loop, with gradients passed through the rounding by the straight-through estimator~\cite{bengio2013estimating}.
Quantizing only after a search on continuous tokens instead lowers PSNR on Mip-NeRF 360 by 0.33\,dB at 8 bits and 0.67\,dB at 6 bits (\cref{subsec:ablations}).
The search runs at the target bit depth, so the code can be refined for any required quantization level.
The target images are renders of the input splat, $\mathbf{Y}_v=\mathcal{R}(\mathcal{G},\mathbf{P}_v)$, which require only the splat and its cameras; when the training images are available, they can serve as targets instead, $\mathbf{Y}_v=\mathbf{I}_v$.
The search changes the code only slightly yet improves fidelity substantially (\cref{subsec:amortization}), which indicates that most of the encoder's remaining error is an amortization gap rather than a limit of what the decoder can express.

\section{Experiments}
\label{sec:experiments}

\subsection{Experimental setup}
\label{subsec:setup}

\paragraph{Evaluation datasets}
On DL3DV~\cite{ling2024dl3dv}, we evaluate on 173 held-out test scenes, which do not overlap with our training scenes.
All methods, including FCGS, compress the same input splats.
Some test scenes may appear in the training set of FCGS, which can only favor FCGS, although our input splats are retrained and differ from those it saw.
To measure generalization to other capture setups and scene sizes, we further evaluate on Mip-NeRF 360~\cite{barron2022mipnerf360}, Tanks and Temples~\cite{knapitsch2017tanks} and Deep Blending~\cite{hedman2018deep}, whose splats contain up to $10\times$ more primitives than the DL3DV training splats.
For these datasets we follow the evaluation protocol of 3DGS~\cite{kerbl20233d} as specified in~\cite{bagdasarian2025survey,morgenstern2026kissgs}.
The held-out views are used neither to fit the input splat nor as targets of the latent search, so all reported metrics are on views that \splattok's search has never seen.

\paragraph{Input splats}
The two main comparisons compress different input splats, so their numbers are not directly comparable.
\Cref{tab:nogt} uses \emph{vanilla 3DGS} splats trained with gsplat~\cite{ye2024gsplat}, whose quality matches the base results reported by FCGS~\cite{chen2025fcgs}, and every method in that table compresses exactly these splats.
\Cref{tab:gt} uses stronger splats trained with \emph{3DGS-MCMC}~\cite{kheradmand2024mcmc}, capped at 3.5 million Gaussians for Mip-NeRF 360, 1.75 million for Tanks and Temples and 1 million for Deep Blending; only \splattok compresses these splats, while the per-scene methods in that table fit their own representation to the training images.
\Cref{tab:gt} is thus an end-to-end comparison in which each method's reconstruction is part of its pipeline; KISS-GS likewise starts from 3DGS-MCMC splats.

\paragraph{Baselines}
Without training images (\cref{tab:nogt}), we compare with the feed-forward codec FCGS~\cite{chen2025fcgs} at its low and high rates, the training-free FlexGaussian~\cite{tian2025flexgaussian}, and LightGaussian~\cite{fan2024lightgaussian} and HAC++~\cite{chen2025hacpp}, which we fine-tune or retrain on renders of the input splat instead of the training images.
All of them except FCGS use the cameras of the input splat, as \splattok's latent search does: FlexGaussian to score the importance of Gaussians, and LightGaussian and HAC++ to render their targets.
With training images (\cref{tab:gt}), we compare with the strongest per-scene methods, ContextGS~\cite{wang2024contextgs} and HAC++~\cite{chen2025hacpp}, which we rerun three times per scene because of its run-to-run variance; with KISS-GS~\cite{morgenstern2026kissgs}, the strongest post-training method, which also decodes to a standard splat; and with representative compact and post-training methods~\cite{papantonakis2024reduced,xie2024mesongs}.
\splattok, the methods in \cref{tab:nogt} and our HAC++ reruns use the same metric implementation and resolution; the remaining numbers in \cref{tab:gt} are taken from the respective papers.

\paragraph{Metrics}
We report PSNR, SSIM and LPIPS (VGG) against the GT images of the held-out test views, the size of the compressed file in MB ($2^{20}$ bytes), and the compression time per scene.
Unless stated otherwise, times exclude fitting the input splat and are measured on a single GH200 GPU.
\splattok's low and high rates are $b{=}6$ and $b{=}8$ bits at one anchor per 128 input Gaussians.
All \splattok results in this report come from one trained model with the same weights, for every dataset, bit depth, anchor density and search target; the network is never retrained or fine-tuned per dataset, rate or scene.

\paragraph{Training data}
We train on 9,101 DL3DV scenes and hold out 173 for testing, every 50th scene of the sorted list, after excluding scenes whose splats fit their training images poorly or whose images have a non-standard resolution.
Each scene's input splat is fit for 10k steps on images at $480{\times}270$ with gsplat.
Each training step renders 32 random views of a scene.

\paragraph{Architecture and training}
Each anchor holds $T{=}16$ tokens of $C{=}32$ channels, the anchors are placed at one per 128 input Gaussians, each anchor takes at most $K{=}640$ members, and the decoder generates $M{=}160$ Gaussians per anchor.
The encoder has width 512 and 12 layers with 8 heads of dimension 64; the first four layers attend within anchors, and the remaining eight alternate between scene-level and within-anchor attention.
The decoder uses one cross-attention layer at the token width, followed by six residual feed-forward layers of width 128 and a linear output head.
Colors use spherical harmonics up to degree $\ell{=}3$ for luma and the DC term only for the two chroma channels.
The model has 57.3M parameters: 56.0M in the encoder (50.4M in its 12 transformer layers, 5.0M in the cross-attention from the learned anchor queries to the Gaussians, and 0.6M in the input projection, query and position embeddings, and the output projection to the tokens) and 1.25M in the decoder.
We train with Muon~\cite{jordan2024muon} for the 86 weight matrices (56.6M parameters, learning rate scaled by 3 and matched to the AdamW update size) and AdamW for the remaining parameters, with weight decay $10^{-4}$, gradient clipping at 1.0, $\lambda{=}0.2$ in \cref{eq:loss}, bf16 mixed precision, and a batch of 16 scenes (including gradient accumulation of 4).
Training runs for 300 epochs at a learning rate of $5{\times}10^{-5}$ after a 10-epoch linear warm-up, with gradient checkpointing and FlashAttention-2, and takes about 6 days on 4 NVIDIA A100 40GB GPUs.

\paragraph{Latent search}
We first search the 8-bit code and then derive the 6-bit code from it.
The 8-bit search starts from the encoder's code and runs 500 steps of Adam on 32 random training views per step, with a peak learning rate of 0.2 decayed to 0.02 with a cosine schedule (on DL3DV, 300 steps with a peak learning rate of 0.1).
The stored code is an exponential moving average of the iterates with a time constant of 5\% of the steps.
For the 6-bit code, the 8-bit result is rounded to 6 bits and refined for 100 further steps at a peak learning rate of 0.03, again with a moving average.
This is faster than a separate 6-bit search and also gives a better code (\cref{subsec:amortization}).

\paragraph{Quantization and entropy coding}
The per-channel steps $\Delta_c$ are set from the encoder's code at each bit depth and stay fixed during the search.
The rANS coder uses 16-bit probabilities and byte-wise renormalization.
Anchor positions are quantized to 10 bits per axis for the Morton order, and the context neighbor $\pi(k)$ is the nearest of the 32 nearest anchors that precedes $k$ in this order.
The neighbor's symbol is binned into at most 64 bins, with one extra bin for anchors without an earlier neighbor, and the frequency tables are counted once, for each bit depth, from the encoder codes of DL3DV training scenes.
They hold one 16-bit entry per symbol for each (token, channel, neighbor bin) context: 16.2\,MB at 8 bits ($16{\times}32{\times}65$ contexts of 255 symbols) and 3.9\,MB at 6 bits ($16{\times}32{\times}64$ contexts of 63 symbols).
The side information stores the anchor positions and sizes and the per-channel steps in 32-bit floats, 16 bytes per anchor and 4 bytes per channel.

\begin{table}[t]
  \centering
  \caption{\textbf{Compression without training images, on vanilla 3DGS splats} (not the 3DGS-MCMC splats of \cref{tab:gt}).
  All methods compress the same input splats, trained with gsplat~\cite{ye2024gsplat}; fine-tuned methods use renders of the input splat as their target, and all methods except FCGS use its cameras.
  Metrics are against the GT images of the held-out views.
  Size in MB ($2^{20}$ bytes), time in seconds per scene excluding splat fitting.
  Per-scene optimization methods are not run on DL3DV because of their cost over 173 scenes.}
  \label{tab:nogt}
  \resizebox{\textwidth}{!}{%
  \setlength{\tabcolsep}{3pt}
  \begin{tabular}{l ccccc ccccc ccccc}
    \toprule
    Dataset & \multicolumn{5}{c}{DL3DV~\cite{ling2024dl3dv}} & \multicolumn{5}{c}{Mip-NeRF 360~\cite{barron2022mipnerf360}} & \multicolumn{5}{c}{Tanks and Temples~\cite{knapitsch2017tanks} + Deep Blending~\cite{hedman2018deep}} \\
    \cmidrule(lr){2-6} \cmidrule(lr){7-11} \cmidrule(lr){12-16}
    Method & PSNR$\uparrow$ & SSIM$\uparrow$ & LPIPS$\downarrow$ & Size$\downarrow$ & Time$\downarrow$
           & PSNR$\uparrow$ & SSIM$\uparrow$ & LPIPS$\downarrow$ & Size$\downarrow$ & Time$\downarrow$
           & PSNR$\uparrow$ & SSIM$\uparrow$ & LPIPS$\downarrow$ & Size$\downarrow$ & Time$\downarrow$ \\
    \midrule
    \multicolumn{16}{l}{\textit{Uncompressed}} \\
    Input splats (vanilla 3DGS) & 30.86 & 0.905 & 0.094 & 82.96 & -- & 27.49 & 0.813 & 0.223 & 749.1 & -- & 26.55 & 0.872 & 0.212 & 551.23 & -- \\
    \midrule
    \multicolumn{16}{l}{\textit{Training-free compression}} \\
    FlexGaussian~\cite{tian2025flexgaussian} & 29.31 & 0.857 & 0.146 & 4.14 & 6 & 26.78 & 0.787 & 0.268 & 38.42 & 19 & 25.66 & 0.833 & 0.241 & 20.21 & 15 \\
    \midrule
    \multicolumn{16}{l}{\textit{Feed-forward compression}} \\
    FCGS (low rate)~\cite{chen2025fcgs}  & 29.21 & 0.854 & 0.148 & 3.28 & 7 & 27.05 & 0.798 & 0.237 & 36.24 & 33 & 26.27 & 0.844 & 0.230 & 23.20 & 18 \\
    FCGS (high rate)~\cite{chen2025fcgs} & 29.94 & 0.875 & 0.132 & 4.69 & 9 & 27.39 & 0.805 & 0.226 & 64.05 & 44 & 26.44 & 0.859 & 0.219 & 41.99 & 22 \\
    \splattok, encoder only (low rate)  & 26.59 & 0.837 & 0.181 & 0.73 & 2.7 & 22.97 & 0.707 & 0.377 & 6.94 & 82 & 24.01 & 0.804 & 0.355 & 5.01 & 28 \\
    \splattok, encoder only (high rate) & 26.81 & 0.842 & 0.194 & 1.08 & 2.7 & 23.05 & 0.714 & 0.371 & 9.85 & 82 & 24.18 & 0.810 & 0.350 & 7.35 & 28 \\
    \midrule
    \multicolumn{16}{l}{\textit{Fine-tuned on or trained using renders of the input splat}} \\
    LightGaussian~\cite{fan2024lightgaussian} & -- & -- & -- & -- & -- & 27.33 & 0.802 & 0.239 & 45.30 & 855 & 26.15 & 0.844 & 0.240 & 36.82 & 578 \\
    HAC++ (low rate)~\cite{chen2025hacpp}  & -- & -- & -- & -- & -- & 26.76 & 0.783 & 0.274 & 7.01 & 1980 & 25.89 & 0.830 & 0.256 & 4.11 & 1214 \\
    HAC++ (high rate)~\cite{chen2025hacpp} & -- & -- & -- & -- & -- & 26.96 & 0.794 & 0.266 & 15.02 & 1989 & 26.01 & 0.839 & 0.248 & 5.76 & 1221 \\
    \splattok (low rate)  & 29.87 & 0.877 & 0.136 & 0.74 & 45 & 27.21 & 0.801 & 0.237 & 6.99 & 757 & 26.34 & 0.849 & 0.228 & 5.04 & 391 \\
    \splattok (high rate) & 30.08 & 0.889 & 0.128 & 1.09 & 39 & 27.35 & 0.805 & 0.229 & 9.93 & 631 & 26.42 & 0.858 & 0.220 & 7.42 & 326 \\
    \bottomrule
  \end{tabular}}
\end{table}

\subsection{Compression without training images}
\label{subsec:nogt}

\Cref{tab:nogt} compares methods that compress a splat without its training images; all are evaluated against the GT images of the held-out views, not against renders of the input splat.
Without the latent search, the \splattok encoder is far below FCGS in quality, by 2.3 to 4.3\,dB at high rate, although its files are already 4 to 6.5 times smaller.
The latent search closes this gap: \splattok matches the PSNR of FCGS at its high rate (within 0.05\,dB on Mip-NeRF 360 and on Tanks and Temples with Deep Blending, and 0.14\,dB higher on DL3DV) with 4 to 6.5 times smaller files, and exceeds FCGS at its low rate by 0.07 to 0.66\,dB with files 4.4 to 5.2 times smaller.
Relative to the input splats, \splattok compresses Mip-NeRF 360 and Tanks and Temples with Deep Blending about 75 and 110 times at its two rates while losing 0.1 to 0.3\,dB.
FlexGaussian and LightGaussian keep the standard format but produce files about 3 to 5 times larger than \splattok at its high rate, at equal or lower PSNR, and LightGaussian needs 10 to 14 minutes per scene.
HAC++, retrained on renders of the input splat, reaches sizes below 10\,MB per scene like \splattok, but with markedly lower quality.

\begin{table}[t]
  \centering
  \caption{\textbf{Comparison with per-scene methods, on 3DGS-MCMC splats} (not the vanilla 3DGS splats of \cref{tab:nogt}).
  All methods start from the same training images and are evaluated on the same held-out views; how each method reconstructs the scene is part of its pipeline.
  HAC++ and ContextGS fit their own representation; KISS-GS and \splattok compress 3DGS-MCMC splats.
  \splattok's latent search targets the training images, or, for \splattok$^\diamond$, renders of the input splat, using no training images.
  HAC++ is the mean of three reruns, matching the re-evaluation by KISS-GS.
  $^\dagger$Reported by the authors.
  Size in MB ($2^{20}$ bytes).}
  \label{tab:gt}
  \small
  \setlength{\tabcolsep}{2.3pt}
  \begin{tabular}{l cccc cccc cccc}
    \toprule
    Dataset & \multicolumn{4}{c}{Mip-NeRF 360~\cite{barron2022mipnerf360}} & \multicolumn{4}{c}{Tanks and Temples~\cite{knapitsch2017tanks}} & \multicolumn{4}{c}{Deep Blending~\cite{hedman2018deep}} \\
    \cmidrule(lr){2-5} \cmidrule(lr){6-9} \cmidrule(lr){10-13}
    Method & PSNR$\uparrow$ & SSIM$\uparrow$ & LPIPS$\downarrow$ & Size$\downarrow$
           & PSNR$\uparrow$ & SSIM$\uparrow$ & LPIPS$\downarrow$ & Size$\downarrow$
           & PSNR$\uparrow$ & SSIM$\uparrow$ & LPIPS$\downarrow$ & Size$\downarrow$ \\
    \midrule
    3DGS$^\dagger$~\cite{kerbl20233d} & 27.46 & 0.812 & 0.222 & 750.9 & 23.69 & 0.844 & 0.178 & 431.0 & 29.42 & 0.899 & 0.247 & 663.9 \\
    Input splats (3DGS-MCMC) & 28.17 & 0.834 & 0.189 & 787.7 & 24.58 & 0.872 & 0.139 & 393.9 & 30.12 & 0.913 & 0.246 & 225.1 \\
    \midrule
    Reduced3DGS$^\dagger$~\cite{papantonakis2024reduced} & 27.19 & 0.807 & 0.230 & 29.54 & 23.57 & 0.840 & 0.188 & 14.00 & 29.63 & 0.902 & 0.249 & 18.00 \\
    MesonGS-FT$^\dagger$~\cite{xie2024mesongs} & 26.99 & 0.796 & 0.247 & 27.16 & 23.32 & 0.837 & 0.193 & 16.99 & 29.51 & 0.901 & 0.251 & 24.76 \\
    ContextGS (low rate)$^\dagger$~\cite{wang2024contextgs} & 27.62 & 0.808 & 0.237 & 12.68 & 24.20 & 0.852 & 0.184 & 7.05 & 30.11 & 0.907 & 0.265 & 3.45 \\
    ContextGS (high rate)$^\dagger$~\cite{wang2024contextgs} & 27.75 & 0.811 & 0.231 & 18.41 & 24.29 & 0.855 & 0.176 & 11.80 & 30.39 & 0.909 & 0.258 & 6.60 \\
    HAC++ (low rate)~\cite{chen2025hacpp} & 27.47 & 0.802 & 0.255 & 8.61 & 24.16 & 0.848 & 0.190 & 5.26 & 30.00 & 0.900 & 0.269 & 2.94 \\
    HAC++ (high rate)~\cite{chen2025hacpp} & 27.60 & 0.806 & 0.233 & 18.48 & 24.33 & 0.853 & 0.178 & 8.62 & 30.19 & 0.905 & 0.259 & 5.31 \\
    \midrule
    KISS-GS$^\dagger$ (low rate)~\cite{morgenstern2026kissgs} & 27.75 & 0.809 & 0.239 & 7.17 & 24.48 & 0.853 & 0.183 & 3.62 & 29.45 & 0.896 & 0.288 & 1.87 \\
    KISS-GS$^\dagger$ (high rate)~\cite{morgenstern2026kissgs} & 27.94 & 0.823 & 0.208 & 12.76 & 24.73 & 0.868 & 0.151 & 7.16 & 29.75 & 0.901 & 0.268 & 3.73 \\
    \midrule
    \splattok$^\diamond$ (low rate) & 27.53 & 0.801 & 0.223 & 7.13 & 24.24 & 0.835 & 0.196 & 3.71 & 30.04 & 0.903 & 0.273 & 2.06 \\
    \splattok$^\diamond$ (high rate) & 27.68 & 0.808 & 0.216 & 10.64 & 24.32 & 0.840 & 0.189 & 5.47 & 30.15 & 0.906 & 0.269 & 3.06 \\
    \splattok (low rate) & 27.93 & 0.813 & 0.222 & 7.46 & 24.67 & 0.847 & 0.183 & 3.94 & 30.18 & 0.903 & 0.271 & 2.08 \\
    \splattok (high rate) & 28.09 & 0.820 & 0.215 & 10.89 & 24.77 & 0.854 & 0.177 & 5.65 & 30.34 & 0.911 & 0.258 & 3.20 \\
    \bottomrule
  \end{tabular}
\end{table}

\subsection{Comparison with per-scene methods}
\label{subsec:gt}

With the training images available, \cref{tab:gt} compares \splattok, searching towards them, with per-scene methods, which all require the training images.
With the training images as target, \splattok outperforms HAC++, the strongest of them, at both rates and on all three datasets, with higher PSNR at smaller sizes.
On Tanks and Temples, it is 0.51 and 0.44\,dB better at 25\% and 34\% smaller sizes, on Deep Blending 0.18 and 0.15\,dB better at 29\% and 40\% smaller sizes, and on Mip-NeRF 360 0.46 and 0.49\,dB better at 13\% and 41\% smaller sizes, with visibly sharper foliage and thin structures (\cref{subsec:qualitative}).
SSIM and LPIPS are comparable to those of HAC++, while KISS-GS has higher SSIM and lower LPIPS at its high rate on Mip-NeRF 360 and Tanks and Temples, with 17 and 27\% larger files.
Even with renders of the input splat as target, and hence without the training images, \splattok stays within 0.04\,dB of HAC++ or exceeds it by up to 0.08\,dB at the same rate, with 17 to 42\% smaller files.
\splattok can even exceed the quality of its own input splat, as on Tanks and Temples, for two reasons: with the training images as the target, the latent search can correct errors of the input splat, and the decoder generates about 1.25 times as many Gaussians as the input contains.
\splattok compresses whatever splat it is given, here splats trained with 3DGS-MCMC~\cite{kheradmand2024mcmc}, so it directly inherits any improvement in splat reconstruction, whereas per-scene methods build their own representation and must be redesigned to benefit from one.
Among the other methods, ContextGS needs 1.7 to 2.1 times larger files for comparable PSNR, KISS-GS reaches similar sizes at up to 0.7\,dB lower PSNR, and the methods that prune and quantize primitives are 3.5 to 12 times larger.
At coarser anchor densities, the margin over HAC++ widens further (\cref{subsec:rate}).

\subsection{Qualitative results}
\label{subsec:qualitative}

\begin{figure}[t]
  \centering
  \includegraphics[width=\textwidth]{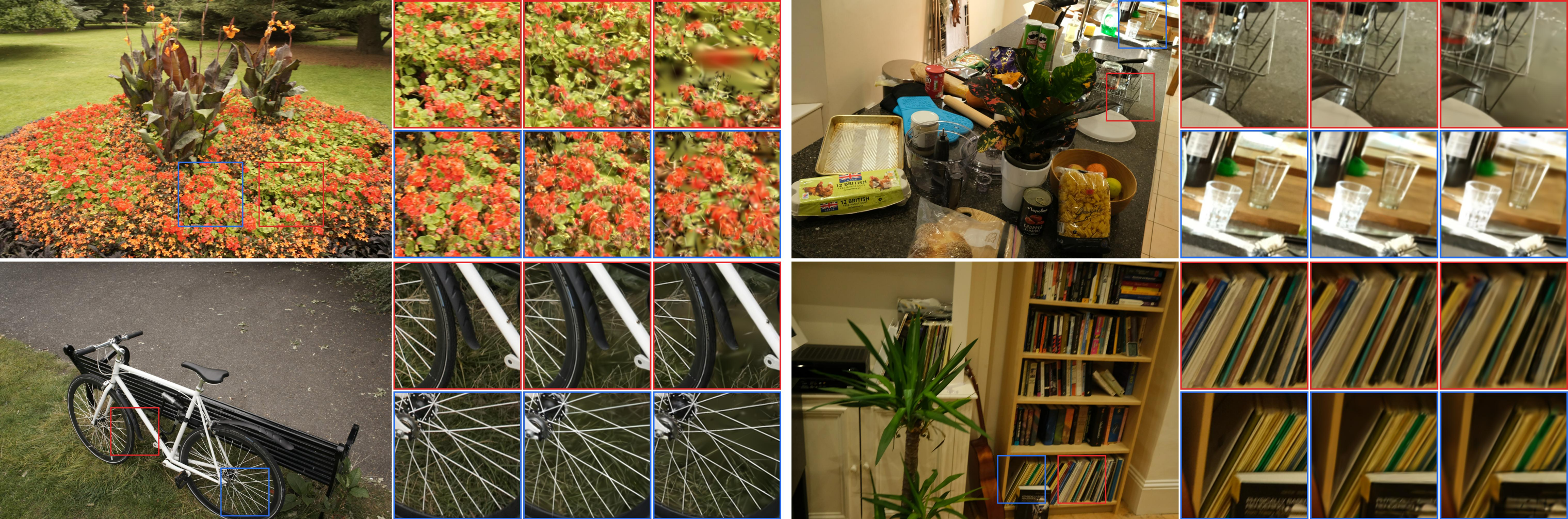}
  \caption{\textbf{Qualitative comparison on Mip-NeRF 360.}
  Left (\textit{flowers} and \textit{bicycle}): high rate. Right (\textit{room} and \textit{counter}): low rate.
  For each set, the leftmost image is the GT image, followed by zoomed-in crops of the GT image, \splattok and HAC++.
  \splattok's latent search targets the training images, as in \cref{tab:gt}.
  \splattok preserves details in foliage, oblique surfaces and thin structures better than HAC++ at both rates.}
  \label{fig:qual}
\end{figure}

\begin{figure}[p]
  \centering
  \includegraphics[width=\textwidth]{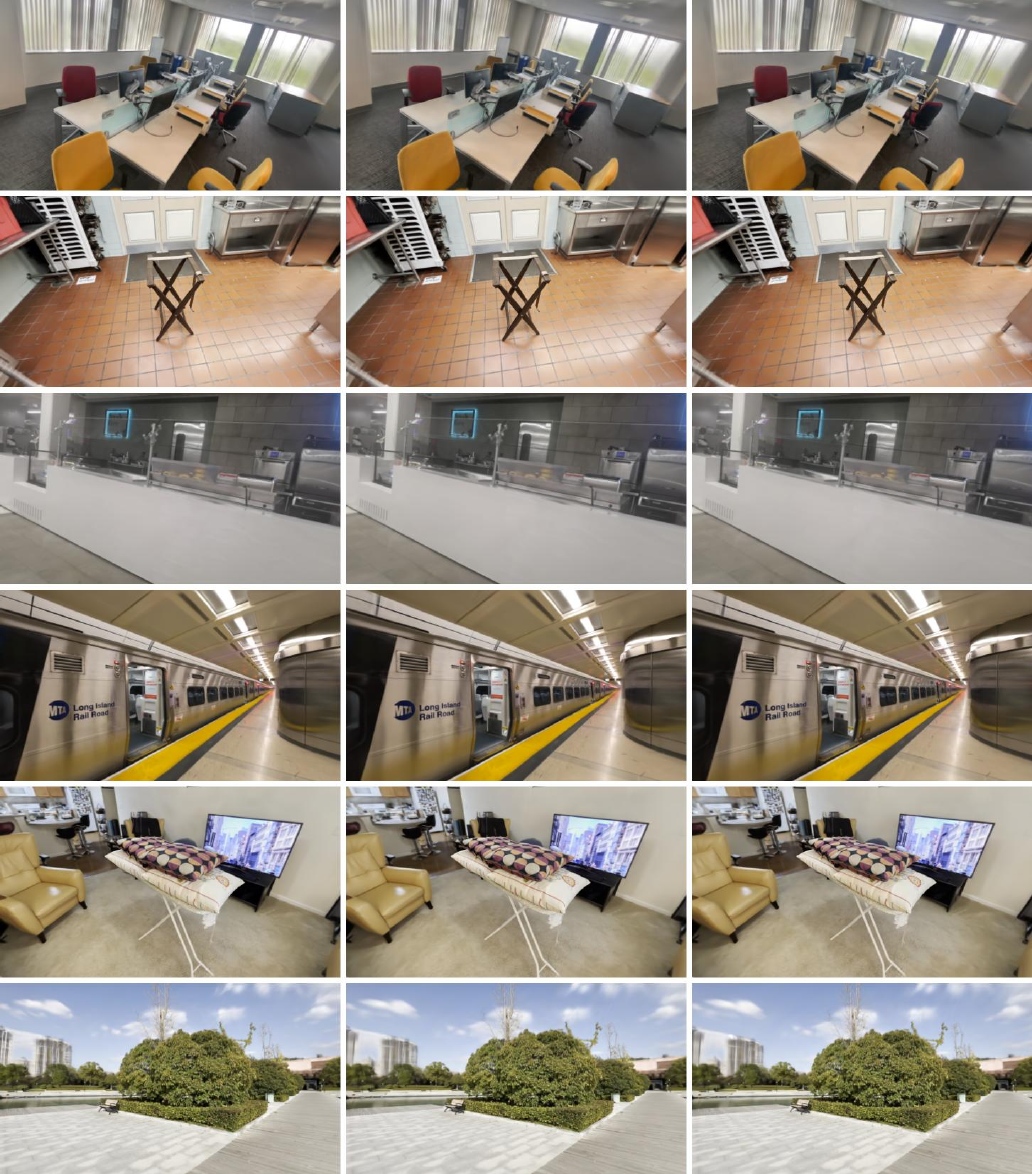}
  \caption{\textbf{DL3DV test scenes.} Left: render of the input splat; middle: \splattok at 8 bits; right: \splattok at 6 bits. The latent search targets renders of the input splat.}
  \label{fig:qual_dl3dv}
\end{figure}

\begin{figure}[p]
  \centering
  \includegraphics[width=\textwidth]{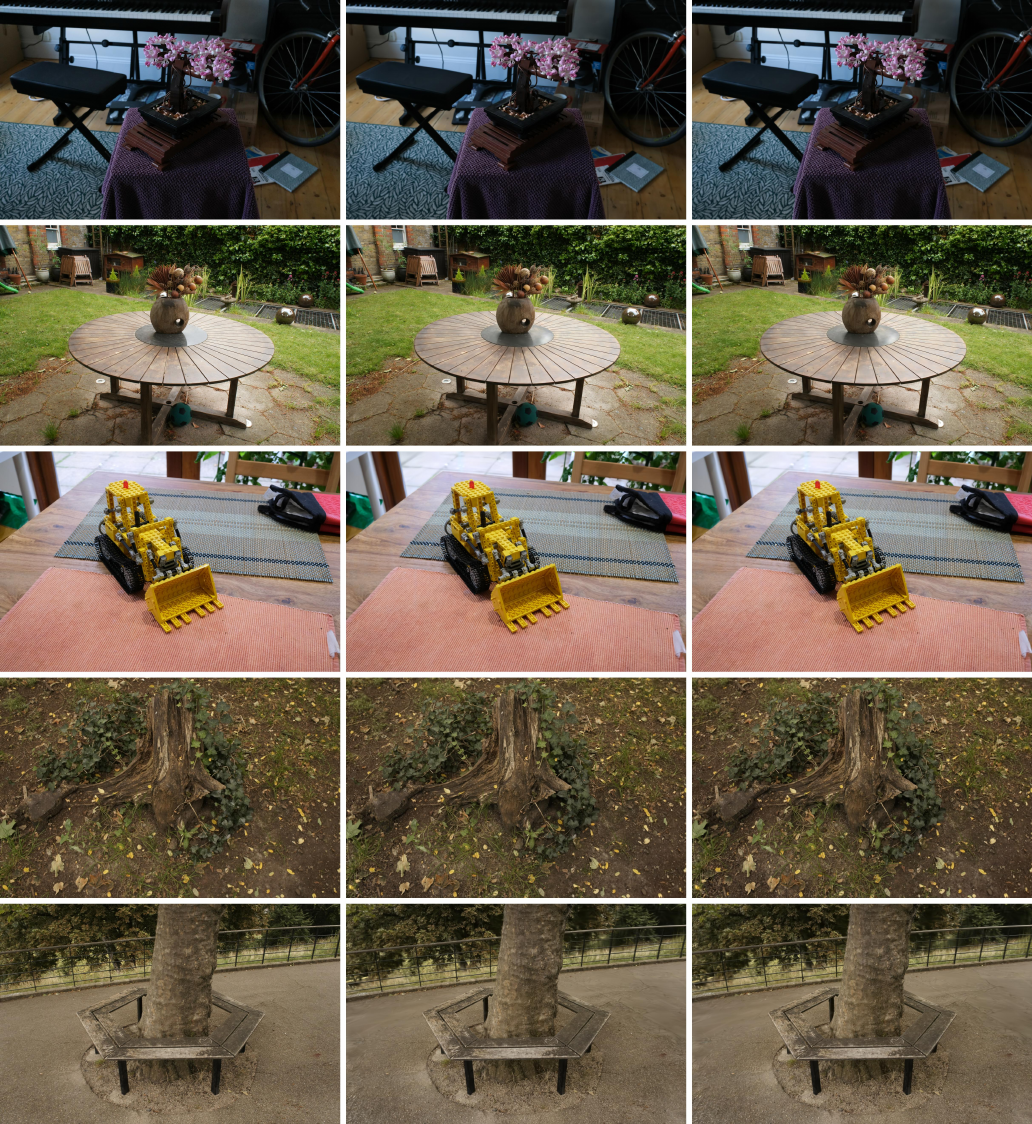}
  \caption{\textbf{Mip-NeRF 360.} Left: GT image; middle: \splattok at 8 bits; right: \splattok at 6 bits. The latent search targets the training images.}
  \label{fig:qual_mip}
\end{figure}

\begin{figure}[p]
  \centering
  \includegraphics[width=\textwidth]{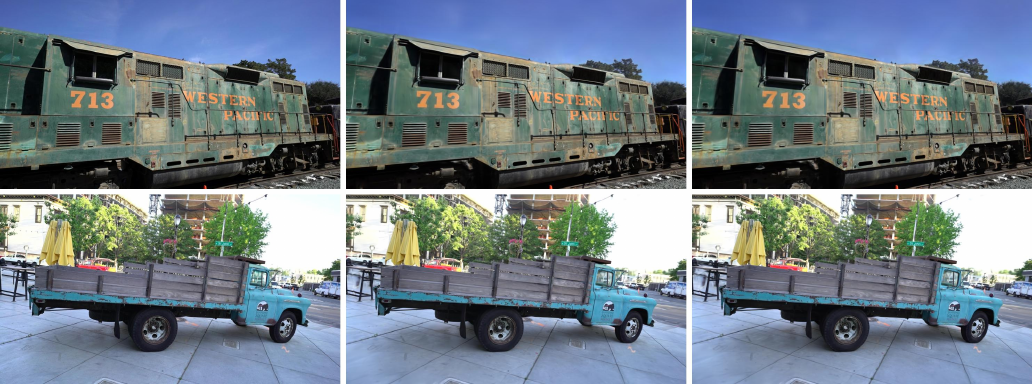}
  \caption{\textbf{Tanks and Temples.} Left: GT image; middle: \splattok at 8 bits; right: \splattok at 6 bits. The latent search targets the training images.}
  \label{fig:qual_tnt}
\end{figure}

\begin{figure}[p]
  \centering
  \includegraphics[width=\textwidth]{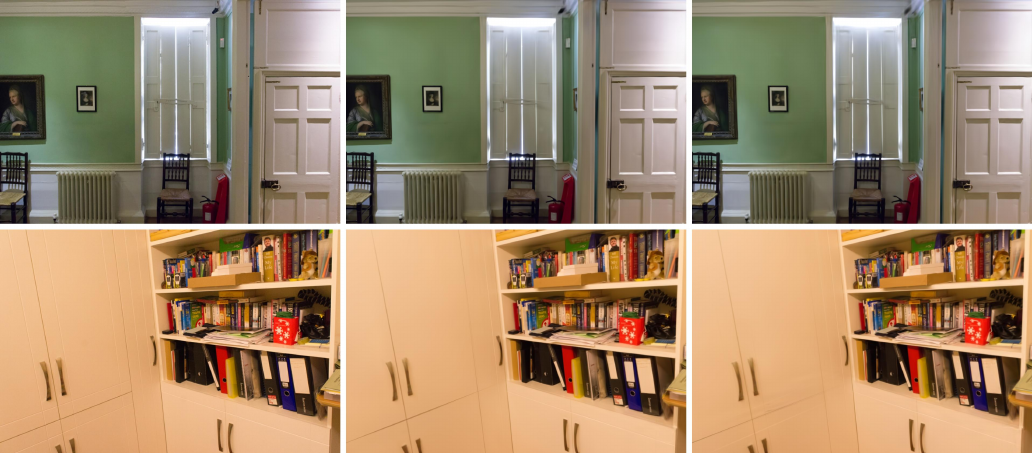}
  \caption{\textbf{Deep Blending.} Left: GT image; middle: \splattok at 8 bits; right: \splattok at 6 bits. The latent search targets the training images.}
  \label{fig:qual_db}
\end{figure}

\Cref{fig:qual} compares \splattok with HAC++ on Mip-NeRF 360 at both rates.
\splattok preserves fine, high-frequency content that HAC++ smooths out, such as foliage, thin structures and surfaces at grazing angles.
\Cref{fig:qual_dl3dv,fig:qual_mip,fig:qual_tnt,fig:qual_db} show further results at both rates on all four datasets: on DL3DV with the latent search towards renders of the input splat, and on Mip-NeRF 360, Tanks and Temples and Deep Blending with the latent search towards the training images.
Going from 8 to 6 bits reduces the file size by about a third with little visible change.

\subsection{Rate control}
\label{subsec:rate}

Since every anchor is coded with the same number of tokens, the size of a code is set by two quantities: the bit depth of the tokens and the number of anchors.
The same trained model serves every setting of both, and since the number of anchors follows the number of input Gaussians, compacting the input splat beforehand acts as a third control.

\begin{table}[t]
  \centering
  \begin{minipage}[t]{0.38\textwidth}
    \centering
    \captionof{table}{\textbf{Bit depth} on \emph{Truck} (Tanks and Temples), with the latent search towards the training images. Bits per value refers to the coded tokens only.}
    \label{tab:bitdepth}
    \footnotesize
    \setlength{\tabcolsep}{3pt}
    \begin{tabular}{l ccc}
      \toprule
      Bit depth & PSNR$\uparrow$ & Size$\downarrow$ & Bits/value$\downarrow$ \\
      \midrule
      $b=4$ & 25.46 & 2.24 & 2.40 \\
      $b=5$ & 26.21 & 3.01 & 3.31 \\
      $b=6$ (low rate) & 26.38 & 3.71 & 4.14 \\
      $b=7$ & 26.49 & 4.59 & 5.18 \\
      $b=8$ (high rate) & 26.51 & 5.47 & 6.23 \\
      \bottomrule
    \end{tabular}
  \end{minipage}\hfill
  \begin{minipage}[t]{0.59\textwidth}
    \centering
    \captionof{table}{\textbf{Anchor density}: one anchor per 128 input Gaussians (as in training) or per 256, on the 3DGS-MCMC splats of \cref{tab:gt}, with the latent search towards the training images. Time in seconds per scene.}
    \label{tab:density}
    \footnotesize
    \setlength{\tabcolsep}{2.5pt}
    \begin{tabular}{l cc cccc c}
      \toprule
      Dataset & G./anchor & Bits & PSNR$\uparrow$ & SSIM$\uparrow$ & LPIPS$\downarrow$ & Size$\downarrow$ & Time$\downarrow$ \\
      \midrule
      Mip-NeRF 360      & 128 & 8 & 28.09 & 0.820 & 0.215 & 10.89 & 890\\
                        & 128 & 6 & 27.93 & 0.813 & 0.222 & 7.46  & 810\\
                        & 256 & 8 & 27.93 & 0.810 & 0.236 & 5.85  & 531\\
                        & 256 & 6 & 27.76 & 0.802 & 0.243 & 4.04  & 559\\
      \midrule
      Tanks and         & 128 & 8 & 24.77 & 0.854 & 0.177 & 5.65  & 347\\
      Temples           & 128 & 6 & 24.67 & 0.847 & 0.183 & 3.94  & 385\\
                        & 256 & 8 & 24.75 & 0.843 & 0.200 & 3.03  & 188\\
                        & 256 & 6 & 24.57 & 0.836 & 0.206 & 2.14  & 202\\
      \midrule
      Deep Blending     & 128 & 8 & 30.34 & 0.911 & 0.258 & 3.20  & 198\\
                        & 128 & 6 & 30.18 & 0.903 & 0.271 & 2.08  & 220\\
                        & 256 & 8 & 30.16 & 0.903 & 0.278 & 1.68  & 106\\
                        & 256 & 6 & 30.07 & 0.901 & 0.281 & 1.19  & 116\\
      \bottomrule
    \end{tabular}
  \end{minipage}
\end{table}

\paragraph{Bit depth}
\Cref{tab:bitdepth} varies the precision of the tokens on \emph{Truck}.
Lowering the precision from 8 to 6 and 5 bits costs only 0.15 and 0.35\,dB, while 4 bits costs 1.1\,dB; the same trained model and latent search cover this whole range.

\paragraph{Anchor density}
\Cref{tab:density} uses the model trained with one anchor per 128 input Gaussians, without retraining, at half the anchor density, so that each anchor summarizes about twice as many Gaussians and the decoder emits about 0.6 times as many Gaussians as the input.
The model handles the coarser anchors well: at 8 bits, one anchor per 256 Gaussians matches or nearly matches the PSNR of one anchor per 128 Gaussians at 6 bits (within 0.02\,dB, and 0.08\,dB higher on Tanks and Temples) with 19 to 23\% smaller files.
At low rates, fewer anchors at a higher precision are therefore a better use of the bits than more anchors at a lower precision.
At 6 bits, the coarser anchors reach 1.2 to 4.0\,MB per scene and still exceed HAC++ at its low rate (\cref{tab:gt}) in PSNR, by 0.29\,dB on Mip-NeRF 360, 0.41\,dB on Tanks and Temples and 0.07\,dB on Deep Blending, with 53 to 60\% smaller files.
With half as many anchors, each step of the latent search decodes half as many Gaussians, which roughly halves the search time.
Fine detail suffers more than overall fidelity, however: with fewer decoded Gaussians per region, SSIM and LPIPS fall further behind the default density at 6 bits than PSNR does.
This suggests varying the anchor density within a scene, so that intricate regions receive more tokens than simple ones (\cref{sec:discussion}).

\begin{table}[t]
  \centering
  \caption{\textbf{Compressing compacted splats} on Mip-NeRF 360.
  The input splats are compacted to 512k Gaussians per scene with POPSpa, the compaction stage of KISS-GS~\cite{morgenstern2026kissgs}, and were shared with us by its authors.
  \splattok compresses them at 8 and 6 bits with the latent search towards the training images, as KISS-GS's fine-tuning does.
  $^\dagger$Read from the rate-distortion plot of KISS-GS (SOG-XT encoding with encoding-aware fine-tuning).
  $^\ddagger$Interpolated linearly in log size between its 64k and 128k operating points, at the sizes of \splattok.
  Size in MB ($2^{20}$ bytes).}
  \label{tab:compacted}
  \small
  \setlength{\tabcolsep}{5pt}
  \begin{tabular}{l ccccc}
    \toprule
    Method & \#Gaussians & PSNR$\uparrow$ & SSIM$\uparrow$ & LPIPS$\downarrow$ & Size$\downarrow$ \\
    \midrule
    POPSpa splats (uncompressed input)           & 512\,k & 27.79 & 0.813 & 0.239 & -- \\
    \midrule
    KISS-GS$^\dagger$, 512k budget               & 512\,k & 27.75 & 0.809 & 0.239 & 7.17 \\
    KISS-GS$^\dagger$, 128k budget               & 128\,k & 26.69 & 0.754 & 0.322 & 1.84 \\
    KISS-GS$^\ddagger$, interpolated             & --     & 26.62 & 0.750 & 0.327 & 1.74 \\
    KISS-GS$^\ddagger$, interpolated             & --     & 26.19 & 0.725 & 0.355 & 1.25 \\
    POPSpa splats + \splattok (8 bits)           & --     & 26.72 & 0.751 & 0.325 & 1.74 \\
    POPSpa splats + \splattok (6 bits)           & --     & 26.50 & 0.730 & 0.332 & 1.25 \\
    \bottomrule
  \end{tabular}
\end{table}

\paragraph{Compacted inputs}
\splattok accepts any standard splat, including splats already reduced by compaction methods, and compacting the input shrinks its bitstream in proportion.
\Cref{tab:compacted} applies \splattok to the 512k-Gaussian POPSpa splats that KISS-GS~\cite{morgenstern2026kissgs} produces before its own encoding.
At 8 bits, \splattok compresses them to 1.74\,MB at 26.72\,dB, on par with KISS-GS at its 128k operating point (26.69\,dB at 1.84\,MB), and 4.1 times smaller than KISS-GS's own encoding of the same 512k splats.
At 6 bits, it reaches 26.50\,dB at 1.25\,MB, above the KISS-GS curve, which interpolates to 26.19\,dB at this size, with higher SSIM and lower LPIPS as well.
\splattok thus matches or exceeds the trade-off of KISS-GS from a less compacted splat, and can serve as the final encoding stage of a compaction pipeline.

\subsection{Latent search and the amortization gap}
\label{subsec:amortization}

\begin{wrapfigure}{r}{0.47\textwidth}
  \vspace{-1.2em}
  \centering
  \includegraphics[width=0.47\textwidth]{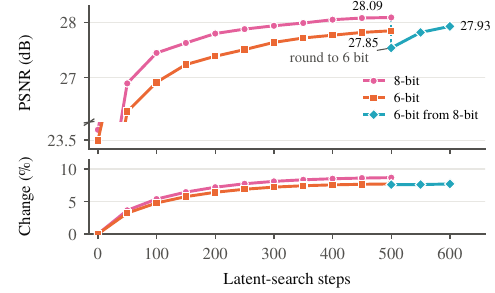}
  \caption{\textbf{Latent search on Mip-NeRF 360}, averaged over its scenes, towards the training images.
  Top: PSNR on the held-out views.
  Bottom: relative root-mean-square change of the latent code from the encoder's output.}
  \label{fig:search}
  \vspace{-1em}
\end{wrapfigure}

\Cref{fig:search} shows how quality and the code evolve during the latent search.
The search improves the encoder's code quickly: on Mip-NeRF 360, the 8-bit code gains 3.7\,dB within the first 100 steps and levels off at 28.09\,dB after 500 steps, while the code moves by less than 10\% from the encoder's output.
Rounding the searched 8-bit code to 6 bits and refining it for 100 more steps gives 27.93\,dB, above a 6-bit search from scratch (27.85\,dB) at a fraction of its cost, and the derived 6-bit code stays within the same range of change.
The search barely affects the rate: the size of the entropy-coded bitstream changes by less than 1\%.

Measured against renders of the input splat on the held-out test views, which the search never uses, the gain is approximately 8\,dB, yet the code changes by only 7 to 8\% in relative root-mean-square terms; across datasets the change ranges from about 4\% on DL3DV to 10\% on Tanks and Temples.
Using the same frozen decoder, the search thus removes 84\% of the encoder's squared error with respect to the input splat.
The best code the decoder admits is at least as good as the searched one, so at least this fraction of the encoder's error is an amortization gap, and at most 16\% can be attributed to the decoder.
Moreover, the better code lies close to the encoder's: the encoder, which must serve all scenes at once, is nearly right for each of them, which explains why a short search with the network frozen suffices.
The two search targets behave similarly, and searching towards the training images adds a further 0.4\,dB, since the GT images also contain detail that the input splat itself does not reproduce.

\paragraph{Does the search need the encoder?}
\begin{wraptable}{r}{0.47\textwidth}
  \vspace{-1.2em}
  \centering
  \caption{\textbf{Entropy coding} on \emph{Truck} at high rate. The variants are lossless and differ only in size.}
  \label{tab:entropy}
  \small
  \setlength{\tabcolsep}{4pt}
  \begin{tabular}{l cc}
    \toprule
    Probability tables & Size$\downarrow$ & Bits/value$\downarrow$ \\
    \midrule
    Single histogram                 & 6.17 & 7.06 \\
    Per position                     & 5.49 & 6.26 \\
    Per position + neighbor context  & 5.47 & 6.23 \\
    \bottomrule
  \end{tabular}
  \vspace{-0.8em}
\end{wraptable}
To test this, we initialize the tokens of \emph{Truck} randomly, from the per-channel mean and variance of the tokens of 50 random DL3DV training scenes, keep the anchors, anchor sizes and quantization steps of the input splat, and run the latent search with the same protocol.
The code starts at 7.25\,dB and reaches only 21.44\,dB after the same number of steps, against 26.51\,dB when starting from the encoder's output.
The frozen decoder alone is thus not enough: the encoder places the search near a good code, and the search closes the remaining gap.

\subsection{Ablations}
\label{subsec:ablations}

\paragraph{Global attention}
To measure the encoder itself, we compare codes without the latent search, which could compensate for a weaker encoder.
Replacing the global attention layers with attention within each anchor lowers the encoder-only PSNR on the DL3DV test scenes by 0.82\,dB at the same size, confirming that scene-level context improves the code beyond what each anchor's own Gaussians provide.
The latent search narrows this gap, but 0.26\,dB remain after the same number of search steps.
A local-only encoder has a cheaper forward pass, but it does not speed up the latent search, the slowest stage, which uses only the frozen decoder.

\paragraph{Quantization in the loop}
Running the latent search on continuous tokens and quantizing only at the end lowers PSNR on Mip-NeRF 360 by 0.33\,dB at 8 bits and 0.67\,dB at 6 bits, because the continuous code that the search finds does not survive rounding, whereas with quantization in the loop the search finds a code that is good after quantization.

\paragraph{Entropy coding}
Selecting tables by position within the anchor saves 11\% of the bits over a single histogram, and the neighbor-anchor context saves only a further 0.5\%, at no change in quality since the coding is lossless (\cref{tab:entropy}).
Counted histograms thus capture little of the dependence between anchors, which leaves room for learned entropy models (\cref{sec:discussion}).

\subsection{Timing and rendering speed}
\label{subsec:timing}

\Cref{tab:timing} breaks down the time of each stage for scenes of increasing size.
The latent search and decoding scale roughly linearly with the number of anchors, and hence with the number of input Gaussians, while clustering and encoding grow faster, the encoding quadratically because of the global attention; on the largest scenes the latent search still accounts for about 80\% of the time, and GPU-accelerated clustering could shorten the remainder further.
Decoding a stored code takes well under a second even for the largest scenes.
The decoder emits 160 Gaussians per anchor, and anchors are placed at one per 128 input Gaussians, so the decoded splat holds about 1.25 times as many Gaussians as the input; its rendering cost is therefore comparable to that of the input splat, at 171 against 184 frames per second on the \emph{Garden} scene of Mip-NeRF 360 (4.4 against 3.5 million Gaussians, one 1600-pixel-wide frame at a time).

\begin{table}[h]
  \centering
  \caption{\textbf{Timing per stage} for scenes of different sizes, in seconds on a single GH200 GPU.
  Clustering partitions the scene into anchors; encoding and decoding are single network passes; the latent search runs for the default number of steps.}
  \label{tab:timing}
  \small
  \setlength{\tabcolsep}{6pt}
  \begin{tabular}{rr ccccc}
    \toprule
    Gaussians & Anchors & Clustering & Encoding & Latent search & Entropy decoding & Decoding \\
    \midrule
    350\,k & 2736  & 3.1 & 0.10 & 48    & 0.05 & 0.04 \\
    1\,M   & 7813  & 16.6 & 0.51 & 179   & 0.13 & 0.11 \\
    3.5\,M & 27559 & 143 & 5.1  & 672  & 0.48 & 0.44 \\
    \bottomrule
  \end{tabular}
\end{table}

Among methods that reach comparable compression ratios, \splattok's compression time is similar or lower.
On the same input splats and GPU, HAC++ retrained on renders of the input splat takes about 33 minutes per scene on Mip-NeRF 360, against 10 to 13 minutes for \splattok (\cref{tab:nogt}).
KISS-GS~\cite{morgenstern2026kissgs} reports timings for a single scene, \emph{Bicycle} from Mip-NeRF 360, on one A100 GPU: at its higher rate, compaction, encoding and encoding-aware fine-tuning take about 950 seconds, excluding the reconstruction of the input splat, against roughly 820 seconds for \splattok on a splat of 3.5 million Gaussians (\cref{tab:timing}).
Since the KISS-GS number is read from the runtime plot in their paper and measured on different hardware, this comparison is approximate.
Feed-forward and training-free methods such as FCGS and FlexGaussian compress a scene in seconds (\cref{tab:nogt}), but produce files 3 to 6.5 times larger.

\section{Discussion}
\label{sec:discussion}

\paragraph{Limitations}
The latent search is the slowest stage: it takes close to three minutes on scenes of one million Gaussians, far longer than a single forward pass of the encoder.
The encoder also needs substantial GPU memory on large scenes, about 30\,GB for four million Gaussians, because its global attention spans all tokens of the scene; a chunked implementation could reduce this.

\paragraph{Accelerating the latent search}
Decoding all anchors is by far the most expensive part of each search step on scenes with many Gaussians.
Since each step renders only a few cameras, decoding only the anchors visible from these cameras would reduce this cost.
Fewer anchors also shorten the search, as \cref{subsec:rate} shows.

\paragraph{Adaptive anchor density}
The anchor-density experiment (\cref{subsec:rate}) shows that the same model handles coarser anchors well, but that fine detail suffers more than overall fidelity when every region receives fewer tokens.
Allocating anchors according to local complexity, rather than uniformly per input Gaussian, would spend the bits where they matter most.

\paragraph{Learned entropy models}
The counted probability tables capture little of the dependence between neighboring anchors (\cref{subsec:ablations}).
A learned entropy model conditioned on already decoded tokens, as in learned image compression~\cite{balle2018variational,minnen2018joint}, could reduce the rate further, and a rate term in the latent search would let it trade size and quality per scene.

\paragraph{Scaling with data}
As a learned codec, \splattok should further benefit from more training data: without the latent search, the encoder scores 1.8\,dB higher on training scenes than on unseen ones, a sign that 9,101 scenes do not yet cover the diversity of real captures, while the latent search closes this gap on unseen scenes.
Better data augmentation for training the autoencoder, for example over scene scale, density and capture style, is a complementary way to reduce this gap.
Since the latent search primarily compensates for errors in the amortized code, a better encoder trained on larger and more diverse data may also reduce the number of search steps needed to reach a given quality.

\paragraph{Scene tokens beyond compression}
Beyond compression, the tokens form a compact, 3D-anchored description of a whole scene, a few thousand anchors instead of millions of primitives, which makes them a natural interface for downstream tasks such as scene generation, editing, retrieval and semantic understanding.

\section{Conclusion}
\label{sec:conclusion}

Instead of coding the primitives of a given splat or optimizing a new representation per scene from scratch, \splattok resynthesizes the scene: it learns a latent space of spatially anchored tokens from thousands of splats and compresses a splat by searching for its code in this space with the networks frozen.
A single model compresses splats from different datasets at a wide range of rates without their training images, exceeds per-scene methods in PSNR at smaller sizes when these images are available, and decodes to a standard splat.
Because it compresses whatever splat it is given, it also inherits future improvements in splat reconstruction.

\section*{Acknowledgments}
This work began when Sai Karthikey Pentapati interned at Dolby Labs. We also thank Texas Advanced Computing Center (TACC) For granting us access to the Vista server. We thank the authors of KISS-GS for sharing their POPSpa-compacted splats.

{\small
\bibliographystyle{ieeenat_fullname}
\bibliography{main}
}

\clearpage
\appendix
\section{Per-scene results}
\label{app:scenewise}

\Cref{tab:scene_mip,tab:scene_tnt,tab:scene_db} list the per-scene results of \splattok on Mip-NeRF 360, Tanks and Temples and Deep Blending, for the 3DGS-MCMC input splats with the latent search towards the training images, as in \cref{tab:gt}.
For each scene we report PSNR, SSIM and LPIPS on the held-out test views at the low and the high rate; the last row of each table gives the mean over the scenes.

\begin{table}[h]
  \centering
  \caption{\textbf{Per-scene results on Mip-NeRF 360} with 3DGS-MCMC input splats and the latent search towards the training images.}
  \label{tab:scene_mip}
  \small
  \setlength{\tabcolsep}{6pt}
  \begin{tabular}{l ccc ccc}
    \toprule
    & \multicolumn{3}{c}{Low rate} & \multicolumn{3}{c}{High rate} \\
    \cmidrule(lr){2-4} \cmidrule(lr){5-7}
    Scene & PSNR$\uparrow$ & SSIM$\uparrow$ & LPIPS$\downarrow$ & PSNR$\uparrow$ & SSIM$\uparrow$ & LPIPS$\downarrow$ \\
    \midrule
    Bicycle & 25.18 & 0.734 & 0.264 & 25.35 & 0.747 & 0.256 \\
    Flowers & 21.88 & 0.598 & 0.351 & 21.95 & 0.605 & 0.346 \\
    Garden & 27.01 & 0.824 & 0.175 & 27.22 & 0.835 & 0.165 \\
    Stump & 26.98 & 0.783 & 0.238 & 27.15 & 0.792 & 0.229 \\
    Treehill & 23.32 & 0.653 & 0.327 & 23.34 & 0.659 & 0.319 \\
    Bonsai & 32.85 & 0.948 & 0.171 & 33.08 & 0.952 & 0.166 \\
    Counter & 29.61 & 0.914 & 0.174 & 29.73 & 0.920 & 0.168 \\
    Kitchen & 31.64 & 0.928 & 0.121 & 31.90 & 0.933 & 0.115 \\
    Room & 32.95 & 0.936 & 0.176 & 33.10 & 0.940 & 0.171 \\
    \midrule
    Mean & 27.93 & 0.813 & 0.222 & 28.09 & 0.820 & 0.215 \\
    \bottomrule
  \end{tabular}
\end{table}

\begin{table}[h]
  \centering
  \begin{minipage}[t]{0.48\textwidth}
    \centering
    \captionof{table}{\textbf{Per-scene results on Tanks and Temples}, as in \cref{tab:scene_mip}.}
    \label{tab:scene_tnt}
    \small
    \setlength{\tabcolsep}{3pt}
    \begin{tabular}{l ccc ccc}
      \toprule
      & \multicolumn{3}{c}{Low rate} & \multicolumn{3}{c}{High rate} \\
      \cmidrule(lr){2-4} \cmidrule(lr){5-7}
      Scene & PSNR & SSIM & LPIPS & PSNR & SSIM & LPIPS \\
      \midrule
      Train & 22.97 & 0.814 & 0.214 & 23.02 & 0.820 & 0.208 \\
      Truck & 26.38 & 0.881 & 0.152 & 26.51 & 0.887 & 0.145 \\
      \midrule
      Mean & 24.67 & 0.847 & 0.183 & 24.76 & 0.854 & 0.177 \\
      \bottomrule
    \end{tabular}
  \end{minipage}\hfill
  \begin{minipage}[t]{0.48\textwidth}
    \centering
    \captionof{table}{\textbf{Per-scene results on Deep Blending}, as in \cref{tab:scene_mip}.}
    \label{tab:scene_db}
    \small
    \setlength{\tabcolsep}{3pt}
    \begin{tabular}{l ccc ccc}
      \toprule
      & \multicolumn{3}{c}{Low rate} & \multicolumn{3}{c}{High rate} \\
      \cmidrule(lr){2-4} \cmidrule(lr){5-7}
      Scene & PSNR & SSIM & LPIPS & PSNR & SSIM & LPIPS \\
      \midrule
      Dr Johnson & 29.78 & 0.897 & 0.275 & 29.82 & 0.906 & 0.263 \\
      Playroom & 30.59 & 0.909 & 0.267 & 30.85 & 0.917 & 0.254 \\
      \midrule
      Mean & 30.18 & 0.903 & 0.271 & 30.33 & 0.911 & 0.258 \\
      \bottomrule
    \end{tabular}
  \end{minipage}
\end{table}

\end{document}